\documentclass{isprs} 
\usepackage{setspace}
\usepackage{graphicx,subfigure}
\usepackage{mwe}
\usepackage{diagbox}
\usepackage{geometry} 
\usepackage{epstopdf}
\usepackage[T1]{fontenc}
\usepackage{floatrow}
\usepackage[]{natbib} 
\usepackage[british]{babel} 
\usepackage[hang]{footmisc}
\usepackage{multirow}
\usepackage[utf8]{inputenc}
\usepackage{amsmath,amssymb}
\usepackage{xcolor}

\usepackage{float}
\floatstyle{plaintop}
\restylefloat{table}

\begin{document}
\title{Lake Ice Monitoring with Webcams and Crowd-Sourced Images\vspace{-0.25em}}
\author{\vspace{-0.125em}
 Rajanie Prabha\textsuperscript{1}, Manu Tom\textsuperscript{2}\thanks{corresponding author} , Mathias Rothermel\textsuperscript{3}, Emmanuel Baltsavias\textsuperscript{2}, Laura Leal-Taixe\textsuperscript{1}, Konrad Schindler\textsuperscript{2}}
\address{\textsuperscript{1 }Dynamic Vision and Learning Group, TU Munich, Germany \\
  \textsuperscript{2 }Photogrammetry and Remote Sensing Group, ETH Zurich, Switzerland\\
  \textsuperscript{3 }nFrames GmbH, Stuttgart, Germany \\
	(prabha.rajanie, leal.taixe)@tum.de,
	 (manu.tom, manos, schindler)@geod.baug.ethz.ch, mathias.rothermel@nframes.com
}
\commission{II, }{} 
\workinggroup{II/6\vspace{-0.5em}} 
\icwg{}   
\abstract{Lake ice is a strong climate indicator and has been
  recognised as part of the Essential Climate Variables (ECV) by the
  Global Climate Observing System (GCOS). The dynamics of freezing and
  thawing, and possible shifts of freezing patterns over time, can
  help in understanding the local and global climate systems.  One way to acquire the spatio-temporal information about lake ice formation,
  independent of clouds, is to analyse webcam images. This paper intends to move towards a universal model for monitoring lake
  ice with freely available webcam data. We demonstrate good
  performance, including the ability to generalise across different
  winters and lakes, with a state-of-the-art Convolutional
  Neural Network (CNN) model for semantic image segmentation,
  \textit{Deeplab v3+}. Moreover, we design a variant of that model,
  termed \textit{Deep-U-Lab}, which predicts sharper, more correct
  segmentation boundaries. We have tested the model's ability to
  generalise with data from multiple camera views and two different
  winters. On average, it achieves Intersection-over-Union (IoU)
  values of $\approx$71\% across different cameras and $\approx$69\%
  across different winters, greatly outperforming prior work.  Going
  even further, we show that the model even achieves 60\% IoU on
  arbitrary images scraped from photo-sharing websites. As part of the
  work, we introduce a new benchmark dataset of webcam images,
  \textit{Photi-LakeIce}, from multiple cameras and two different
  winters, along with pixel-wise ground truth annotations.}

\maketitle
\section{INTRODUCTION}\label{introduction}

Climate change is and will continue to be, a main challenge for
humanity. In the words of Stephen Haddrill (2014), "Climate change is
a reality that is happening now, and that we can see its impact
across the world".
Lakes play an essential role in the quest to monitor and better understand the climate system. 
One important piece of information about
lakes in cooler climate zones are the times, duration and patterns of
freezing and thawing. Long-term changes and shifts of these variables
mirror changes in the local climate. Therefore, there is a need to
analyse the temporal dynamics of lake ice, and in fact, it has been
designated an ECV by the GCOS.

This work explores the potential of webcam images, in conjunction with
modern semantic segmentation algorithms such as \textit{Deeplab v3+}
\citep{Deeplabv3plus2018}, for lake ice monitoring. The goal is to
construct a spatially resolved time series of the spatio-temporal
extent of lake ice (note that coarser indicators, e.g., the
\textit{ice-on} and \textit{ice-off} dates, can easily be derived from
the time series).
Given the promising results of \textit{Deeplab v3+} on other
semantic segmentation tasks such as PASCAL VOC \citep{Everingham15}
and Cityscapes \citep{Cordts_2016_CVPR}, we base our approach on that
model.

\begin{figure}[]
\small \subfigure[Webcam RGB image]{\includegraphics[height=1.4cm,
    width=3.4cm]{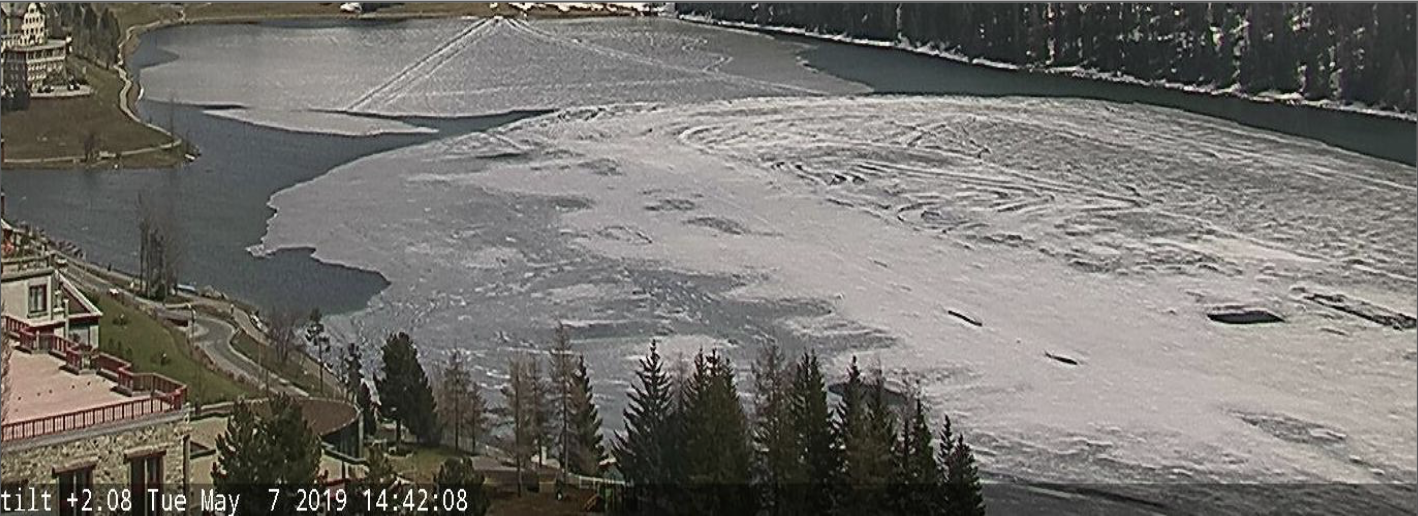}}\hspace{0.1cm} \subfigure[Lake
  detection]{\includegraphics[height=1.4cm,width=3.4cm]{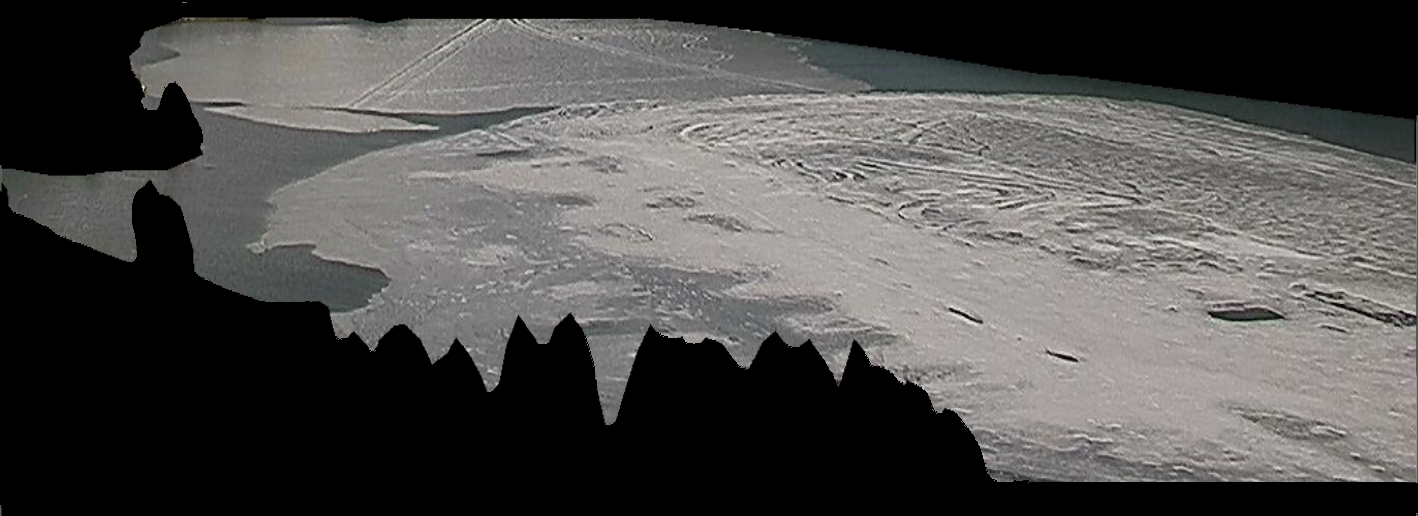}}\\\vspace{-1em} \subfigure[Lake
  ice
  segmentation]{\includegraphics[height=1.4cm,width=3.4cm]{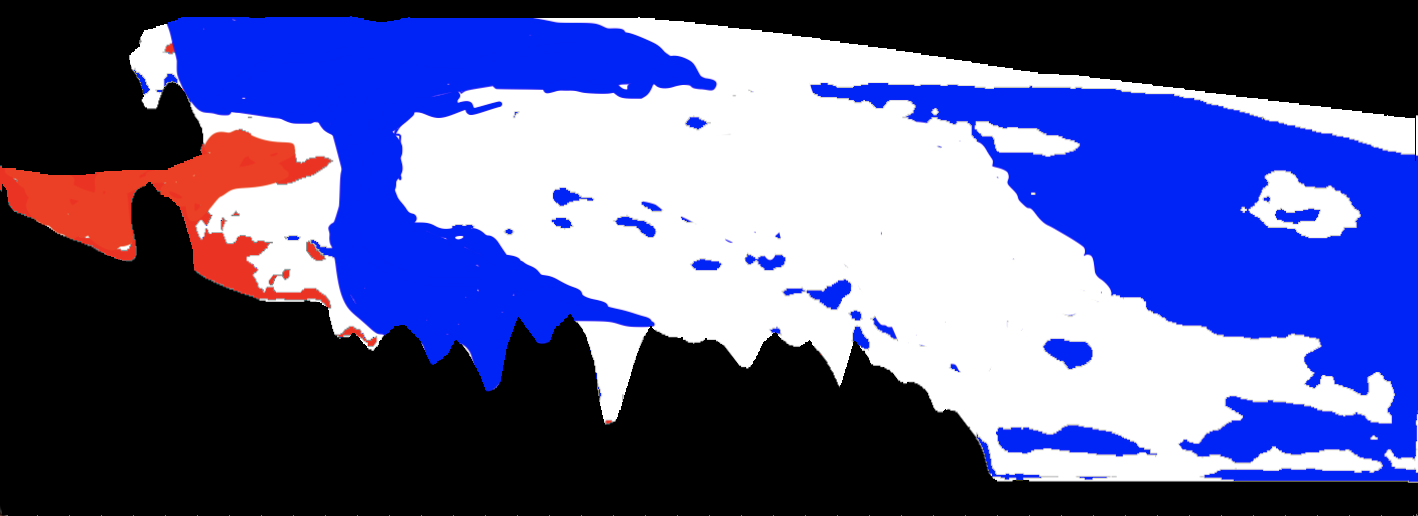}}\hspace{0.1cm}
\subfigure[Ground
  truth]{\includegraphics[height=1.4cm,width=3.4cm]{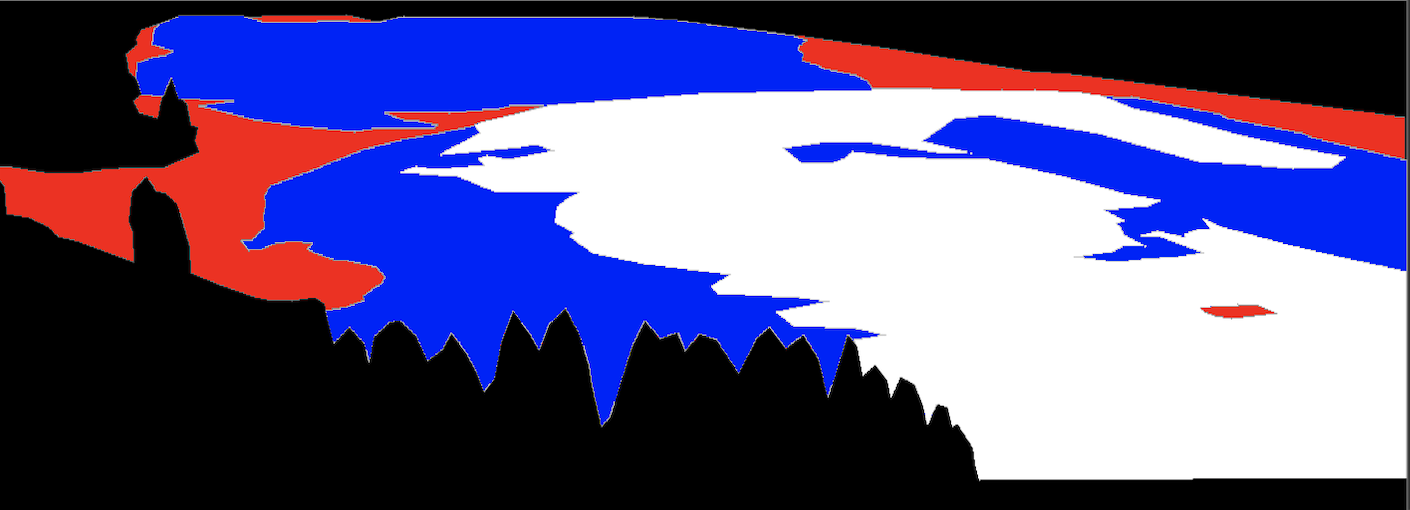}}\vspace{-0.5em}
\subfigure[Colour
  code.]{\includegraphics[width=0.7\columnwidth]{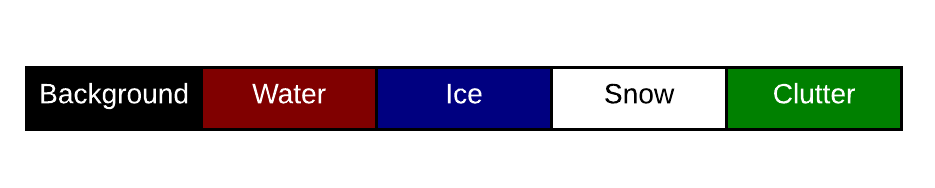}}
   \vspace{-1em}
   \caption{(a) Example webcam image of lake St.~Moritz, from the \textit{Photi-LakeIce} dataset, (b) lake detection result, (c) lake ice segmentation result, (d) corresponding ground truth labels and (e) the colour code used throughout the paper.}
  \label{fig:title_figure}
\end{figure}

The core task for the envisaged monitoring system is: in every camera
frame, classify each pixel capturing the lake surface as
\emph{water}, \emph{ice}, \emph{snow} and \emph{clutter}, i.e., other
objects on the lake, mostly due to human activity such as tents, boats etc. See
Fig.~\ref{fig:title_figure}c.
With a view towards a future operational system, we do lake detection, followed by fine-grained classification. See Fig.~\ref{fig:title_figure}b.
In both steps we take advantage of transfer learning and employ models pre-trained on external databases (here, the PASCAL VOC dataset), to compensate for the relative scarcity of annotated data.

\normalsize

\begin{figure}[ht]
\small
  \subfigure[water]{\includegraphics[height=1.4cm,width=1.8cm]{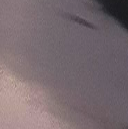}}\hspace{0.3cm} 
  \subfigure[water]{\includegraphics[height=1.4cm,width=1.8cm]{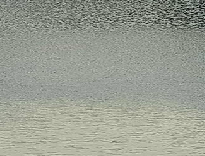}}\hspace{0.3cm}
  \subfigure[water + ice]{\includegraphics[height=1.4cm,width=1.8cm]{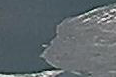}}\\
  \subfigure[ice]{\includegraphics[height=1.4cm,width=1.8cm]{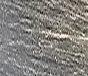}}\hspace{0.3cm}
  \subfigure[snow + ice]{\includegraphics[height=1.4cm,width=1.8cm]{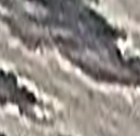}}\hspace{0.3cm}  \subfigure[snow]{\includegraphics[height=1.4cm,width=1.8cm]{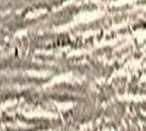}}
   \vspace{-1em}
   \caption{Texture variability of water, ice, and snow in the \textit{Photi-LakeIce} dataset.}
  \label{fig:textures}
\end{figure}

To evaluate any model's ability to generalise, and in particular to
work with high-capacity deep learning methods, one requires a large
and diverse pool of annotated data, i.e., images with pixel-accurate
labels.
Webcams on lakes are a challenging outdoor scenario with limited image
quality, and prone to unfavorable illumination, haze, etc; making it
at times hard to distinguish between ice/snow or water, even for
the human eye, see Fig. \ref{fig:textures}.
For our study, we gathered and annotated several webcam
streams. These include the data from four lakes and three summers for lake detection, and two lakes and two winters for lake ice
segmentation. Entire data is curated and labelled by human
annotators.

\textbf{Contributions.}
\vspace{-0.5em}
\begin{enumerate}
  \setlength\itemsep{0pt}
\item We set a new state of the art for lake ice detection from webcam
  data.
\item Unlike prior art~\citep{prs_report}, our method generalises well across different cameras and lakes, and across different winters.
\item Along the way we also demonstrate automated lake detection; a
  small extension that, however, may be very useful when scaling to many lakes or moving to non-stationary (pan-tilt-zoom) cameras.
\item We introduce \textit{Deep-U-Lab} which produces visibly more accurate
  segment boundaries.
\item We report, for the first time, lake ice detection results for
  crowd-sourced images from image-sharing websites.
\item We make available a new \textit{Photi-LakeIce} dataset of webcam images, with ground truth annotations for multiple lakes and winters.
\end{enumerate}

%
%

\normalsize

\section{RELATED WORK}\label{relatedwork}

\textbf{Lake ice monitoring.} To our knowledge, \cite{muyan_lakeice} proposed lake ice detection with webcams for the first time. The authors used the FC-DenseNet model
\citep{DBLP:journals/corr/JegouDVRB16} and performed experiments on a
single lake (St.~Moritz) for the winter
$2016$-$17$. 
Another work was reported on monitoring lake ice and freezing trends from
low-resolution optical satellite
data~\citep{isprs-annals-IV-2-279-2018}. They used support vector
machines to detect ice and snow on four Alpine lakes in Switzerland 
(Sihl, Sils, Silvaplana, and St.~Moritz). Building on those works,
an integrated monitoring system combining satellite imagery, webcams
and in-situ data was proposed in \cite{prs_report}. Note that this work reported results on two winters ($2016$-$17$ and $2017$-$18$) for the webcam at lake St.~Moritz.
\citeauthor{8900650} and Wang (2019) provided algorithms to generate a bedfast/floating
lake ice product from Synthetic Aperture Radar (SAR), and
\cite{rs10111727} investigated the performance of a semi-automated
segmentation algorithm for lake ice classification using
dual-polarized RADARSAT-2 imagery. \cite{rs11161952} summarised the
physical principles and methods in remote sensing of selected key
variables related to ice, snow, permafrost, water bodies, and
vegetation. 

The starting point for the present work was the observation that the
work of \cite{prs_report} failed to generalise across different
cameras viewing the same lake.  Our goal was to make progress towards
a system that can be applied not only to different views of the same
lake, but also to other lakes and/or data from different winters.
As an even more extreme test, we also test on crowd-sourced data.

\textbf{Amateur images for environmental monitoring.} Besides lake
ice, there are many more domains where images from webcams or
photo-sharing repositories could benefit environmental monitoring.
Examples include \cite{fogmonitor, gasemission, coastal, rs9050408, rs70506133, NorouzzadehE5716, ALBERTON201782, 10.1371/journal.pone.0171918}. Perhaps the closest ones to our work are, on the one
hand, \cite{Salvatori}, where the goal was to detect the extent of snow cover in webcam images; and on the other hand, \cite{DBLP:journals/corr/abs-1901-04412}, where different types of floating ice on rivers were detected with the help of UAV images. We note that crowd-sourcing techniques are, in general, becoming more popular for environmental monitoring, e.g., \cite{hess-20-5049-2016}.

%

\textbf{\textit{Deeplab v3+} for semantic segmentation.} Due to their
unmatched versatility and empirical performance, neural
networks have become the preferred tool for many complex image
analysis tasks, and remote sensing is no
exception.
For the task of semantic segmentation, \textit{Deeplab v3+}
\citep{Deeplabv3plus2018} is one of the most popular architectures,
and the top performer on several different datasets; including generic
consumer pictures, e.g., PASCAL VOC~\citep{Everingham15}, but also
more specific ones like the recent ModaNet
\citep{DBLP:journals/corr/abs-1807-01394}, a large collection of
street fashion images. 
Also in medical image analysis, \textit{Deeplab v3+} has been used to
segment clinical image data, e.g., lesions of the liver in abdominal
CT images \citep{8764548}.
Remote sensing examples include detection of oil spills in satellite
images \citep{Krestenitis_2019} to combat illegal discharges and tank
cleaning that pollute the oceans.
And, closer to our work, detecting different types of ice in UAV
images \citep{DBLP:journals/corr/abs-1901-04412} as an intermediate step
to quantify river ice concentration with relatively small (in deep learning terms) datasets.


\begin{figure*}[ht]
\includegraphics[width=12cm,height=5.5cm]{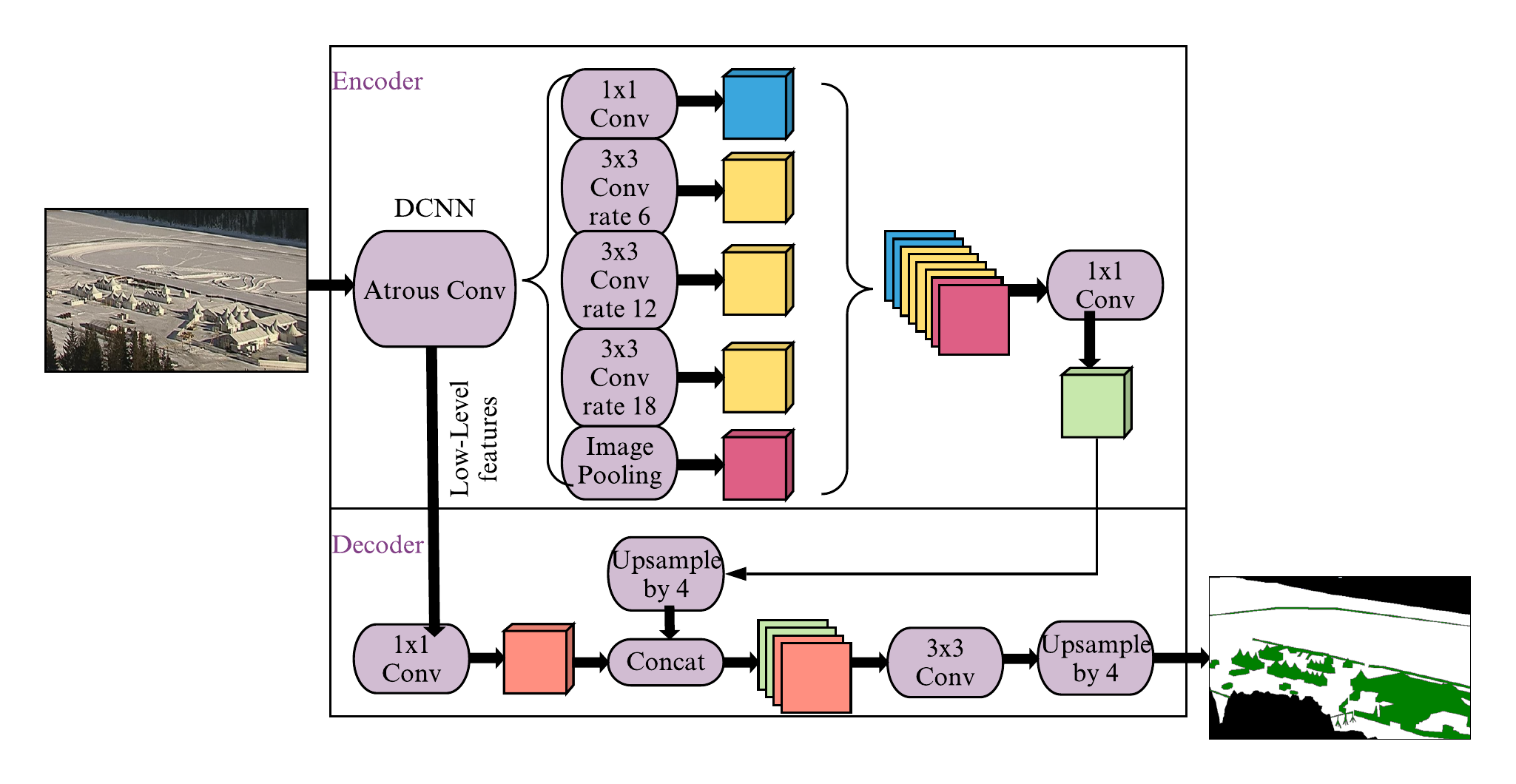}
\vspace{-1.5em}
    \caption{\textit{Deeplab v3+} architecture. Best if viewed on screen.}
    \label{fig:deeplab}
\end{figure*}

\section{Methodology}
\subsection{Deeplab v3+}\label{sec:deeplab v3+}
\textit{Deeplab v3+} \citep{Deeplabv3plus2018} is a CNN architecture
for semantic segmentation, designed to
learn multi-scale contextual features while controlling signal
decimation, see Fig.~\ref{fig:deeplab}. The basic structure is a classical encoder-decoder
architecture. We use \textit{Xception65} as the encoder backbone,
which is similar to the well-known Inception network
\citep{SzegedyLJSRAEVR14}, except that it uses depth-wise separable
convolutions. That is, 2D convolutions are applied on each input channel independently, then combined with 1D convolutions across
channels. This saves a lot of unknowns, without any noticeable
performance penalty. Moreover, all max-pooling operations are replaced
by (depth-wise separable) strided convolutions.

Specific to \textit{Deeplab v3+} is the use of Atrous Spatial Pyramid
Pooling (ASPP), to mitigate spatial smoothing but still encode
multi-scale context. Atrous convolution dilates the kernel by an
integer dilation rate $k$, such that only every $k$-th pixel of the
input layer is used, thus increasing the receptive field without
downsampling the original input.
Overall, the encoder has an output stride (spatial downsampling from
input to final feature encoding) of $16$. In the decoder module, the
encoded features are first upsampled by a factor of $4$, then
concatenated with the low-level features from the corresponding
encoder layer (after reducing the dimensionality of the latter via
$1\times1$ convolution). These resulting ``mid-resolution'' features
are transformed with a further stage of $3\times3$ convolutions, then
upsampled again by a factor $4$ to recover an output map at the full
input resolution.

\begin{figure}[ht]
  \includegraphics[height=2.5cm,width=0.99\textwidth]{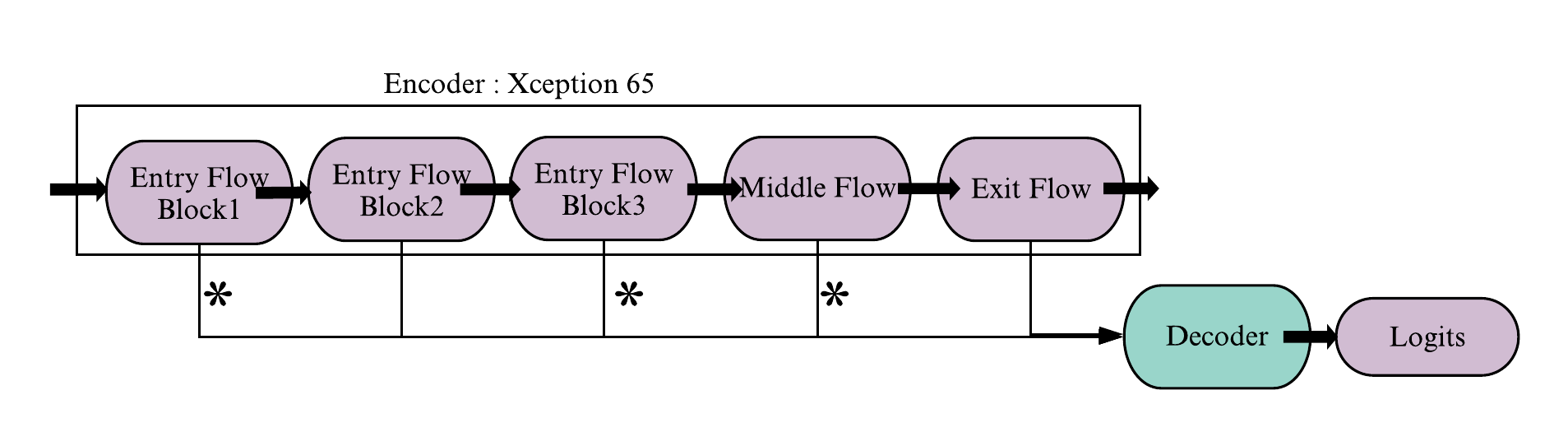}
  \centering
  \caption{\textit{Deep-U-Lab}. The newly added skip connections are marked by "*". Best if viewed on screen.}
  \vspace{-1em}
  \label{fig:deepulab}
\end{figure}

\textbf{Deep-U-Lab.} To mitigate the model's tendency towards overly
smooth, imprecise segment boundaries, we add three extra skip
connections from the entry and middle blocks of the encoder, in the
spirit of \textit{U-net} \citep{RonnebergerFB15}. We call this new
version \textit{Deep-U-Lab}, see Fig.~\ref{fig:deepulab}. The
corresponding feature maps are directly concatenated together with the final output of the encoder block. We found that they help to better preserve high-frequency detail at segment boundaries. The main task of the encoder is to extract high-level features for various classes, with a tendency to loose low-level information not crucial for that task. Hence, we enforce preservation of low-level features through concatenation, so as to refine the class boundaries.  

%
%

\textbf{Transfer learning.} A remarkable property of deep machine
learning models is their ability to learn features that transfer well
across datasets.
We therefore initialise our training with network weights pre-trained
on PASCAL VOC 2012 \citep{Everingham15}, a standardised image dataset for basic objects like animals, people, vehicles, etc..
Even if there seemingly is a considerable domain shift between an
existing image collection (in our case PASCAL) and a new dataset (our
lake ice images), starting from a network learnt for the older
dataset and fine-tuning it quickly adapts it to the  new data and
task, with much less data.
In particular, batch normalization layers for a large network are difficult to train, because it calls for big batch sizes and thus GPU memory. Transfer learning comes to rescue in scenarios like this.
%
\begin{figure*}[ht]
  \small
  \subfigure[Cam0 St.~Moritz]{\includegraphics[height=1.5cm,width=4cm]{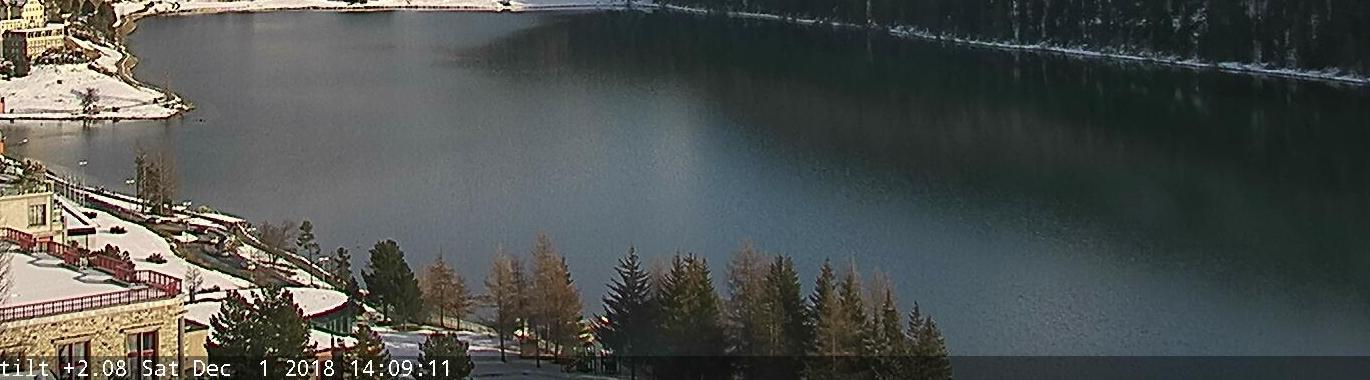}}\hspace{0.1cm}
  \subfigure[Cam0 St.~Moritz FG]{\includegraphics[height=1.5cm,width=4cm]{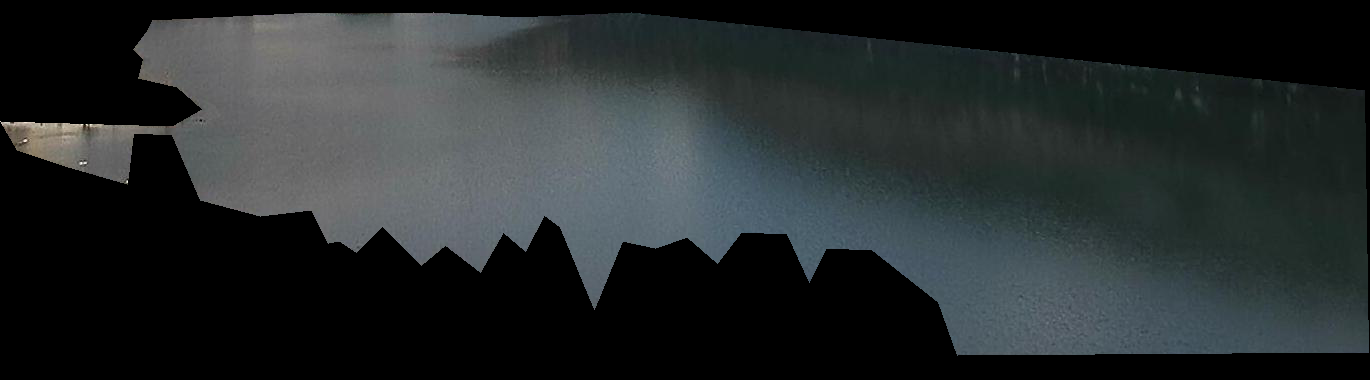}}\hspace{0.1cm}
  \subfigure[Cam1 St.~Moritz]{\includegraphics[height=1.5cm,width=4cm]{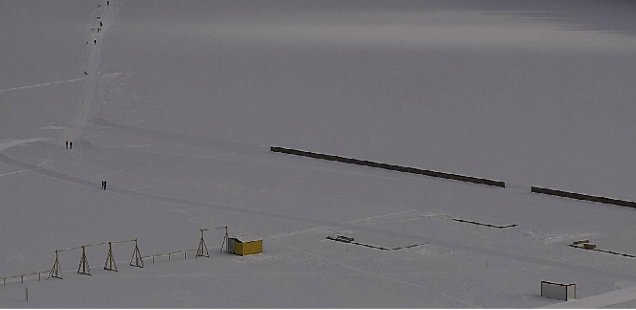}}\hspace{0.1cm}
  \subfigure[Cam1 St.~Moritz FG]{\includegraphics[height=1.5cm,width=4cm]{figures/exampledata_2.png}}\\
  \subfigure[Cam2 Sihl (R1)]{\includegraphics[height=1.5cm,width=4cm]{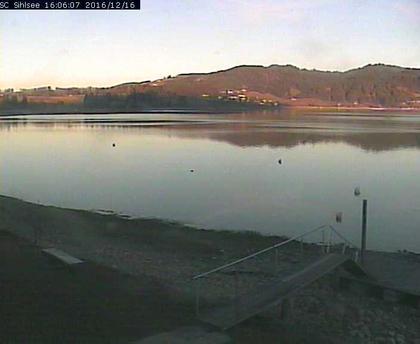}}\hspace{0.1cm}
  \subfigure[Cam2 Sihl (R1) FG]{\includegraphics[height=1.5cm,width=4cm]{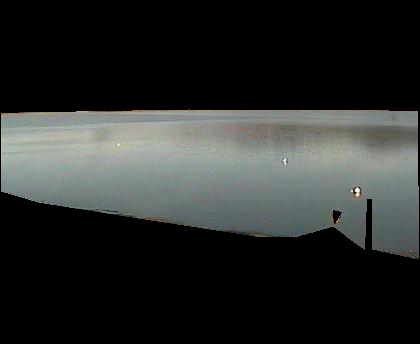}}\hspace{0.1cm}
  \subfigure[Cam2 Sihl (R2)]{\includegraphics[height=1.5cm,width=4cm]{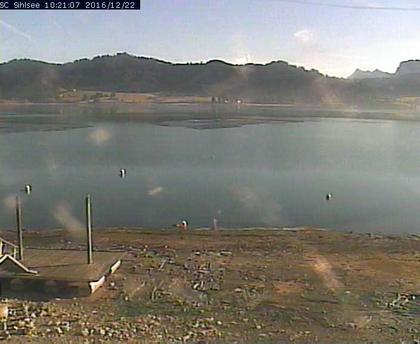}}\hspace{0.1cm}
  \subfigure[Cam2 Sihl (R2) FG]{\includegraphics[height=1.5cm,width=4cm]{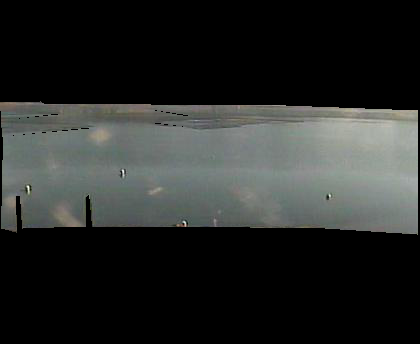}}\\
  \subfigure[Cam2 Sihl (R3)]{\includegraphics[height=1.5cm,width=4cm]{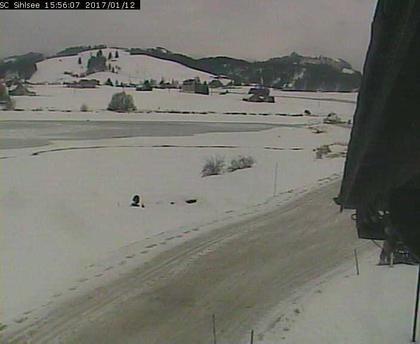}}\hspace{0.1cm}
  \subfigure[Cam2 Sihl (R3) FG]{\includegraphics[height=1.5cm,width=4cm]{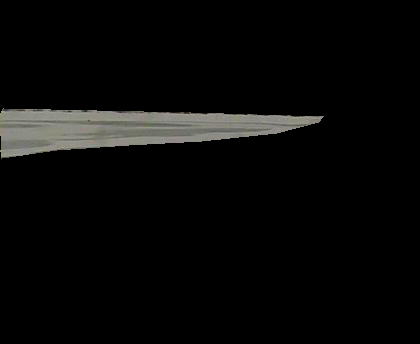}}\hspace{0.1cm}
  \subfigure[Cam2 Sihl (R4)]{\includegraphics[height=1.5cm,width=4cm]{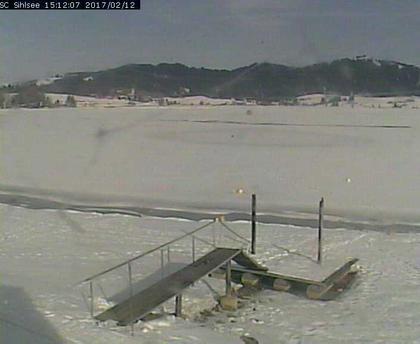}}\hspace{0.1cm}
  \subfigure[Cam2 Sihl (R4) FG]{\includegraphics[height=1.5cm,width=4cm]{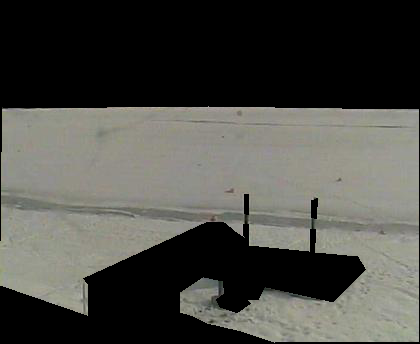}}
  \vspace{-1em}
  \caption{Example images from the \textit{Photi-LakeIce}
    dataset. 1$^\text{st}$ row: fixed cameras monitoring lake
    St.~Moritz. 2$^\text{nd}$ and 3$^\text{rd}$ rows: rotating camera
    (R1, R2, R3, R4 represents different rotations) monitoring lake
    Sihl. Even columns show the foreground (FG) lake area for the
    images shown in the previous column.}
  \label{fig:Photi-lakeice-dataset-snapshots}
\end{figure*}

\normalsize%
\subsection{Lake detection}
It is obvious that classifying lake ice is a lot easier if restricted
to pixels on the lake. Full webcam frames usually include a lot of
background (buildings, mountains, sky, etc.), and passing them
directly to the lake ice classifier can add unnecessary distractions to
the learning and inference stages (e.g., clouds can be difficult to
discriminate from snow).
Therefore, we prefer to localise the lake in a pre-processing step and
run the actual lake ice detection only on lake pixels.
For static webcams, it is relatively easy to localise the lake
manually, as in earlier works ~\citep{muyan_lakeice, prs_report}. There are,
however, situations where an automatic procedure would be preferable,
for instance, if the lake level varies greatly over the
years. Automatic detection of the lake becomes vital if also
crowd-sourced images have to be analysed, since these are typically taken from
variable, unknown viewpoints.

In the context of our work, it is natural to also cast the automatic
lake detection as a two-class (\textit{foreground, background})
pixel-wise semantic segmentation problem and train another instance of the segmentation model. For static webcams, we run the lake detector on
summer images, to sidestep the situation where both the lake
and the surrounding ground is covered with snow.


\subsection{Lake ice segmentation}
\label{lake-ice-segm}

Once the lake mask has been determined, the state of the lake is
inferred with a fine-grained classifier.  In this step, pixels are
labelled as one of four classes (\textit{water, ice, snow,
  clutter}). From the per-pixel maps, we also extract two parameters
often used to describe the temporal dynamics of the freezing cycle:
the \textit{ice-on} date, defined as the first day on which the large
majority of the lake surface is frozen, and which is followed by a
second day with also mostly frozen lake~ (\citeauthor{franssen_scherrer2008} and Scherrer, 2008);
and the \textit{ice-off} date, defined symmetrically as the first day
on which a non-negligible part of the lake surface is liquid water,
and followed by a second non-frozen day.
%
%

\section{Data}

\subsection{Webcam data}
\label{section:webcam_data}

All our webcam images are manually annotated with the
\textit{LabelMe} tool \citep{labelme2016} to generate pixel-wise
ground truth. Additionally, the dataset is cleaned by discarding
excessively noisy images due to bad weather (thick fog, heavy rain, and
extreme illumination conditions). The images vary in spatial resolution,
magnification, and tilt, depending on camera type (fixed or rotating)
and parameters.

\textbf{Lake detection dataset.} For the task of lake detection, we
have collected image streams from four different lakes: one camera each
for lakes Sihl (rotating), Sils (fixed), and St.~Moritz (rotating) and
four cameras (all fixed) for lake Silvaplana. Refer to Table
\ref{tab:lakedetection} for more details.
\begin{table}[ht]
\centering
\small
    \vspace{-0.5em}
    \begin{tabular}{|l|c|c|c|c|}
        \hline
        \textbf{Winter} & \textbf{Lake} & \textbf{Cam} & \textbf{\#images} & \textbf{Res}\\
        \hline
    2016-17 & St.\ Moritz & Cam0 & 820 & 324$\times$1209\\
    2016-17 & St.\ Moritz & Cam1 & 1180 & 324$\times$1209\\
    2016-17 & Sihl & Cam2 & 500 & 344$\times$420\\
    \hline
    2017-18 & St.\ Moritz & Cam0 & 474 & 324$\times$1209\\
    2017-18 & St.\ Moritz & Cam1 & 443 & 324$\times$1209\\
    2017-18 & Sihl & Cam2 & 600 & 344$\times$420\\
\hline
  \end{tabular}
  \caption[]{Key figures of the \textit{Photi-LakeIce} dataset for
    different winters, lakes, and cameras.}
  \label{tab:datasetlakeice}
\end{table}
\normalsize

\begin{figure}[ht]
 \centering
  \fbox{\includegraphics[height=3.5cm,width=7.5cm]{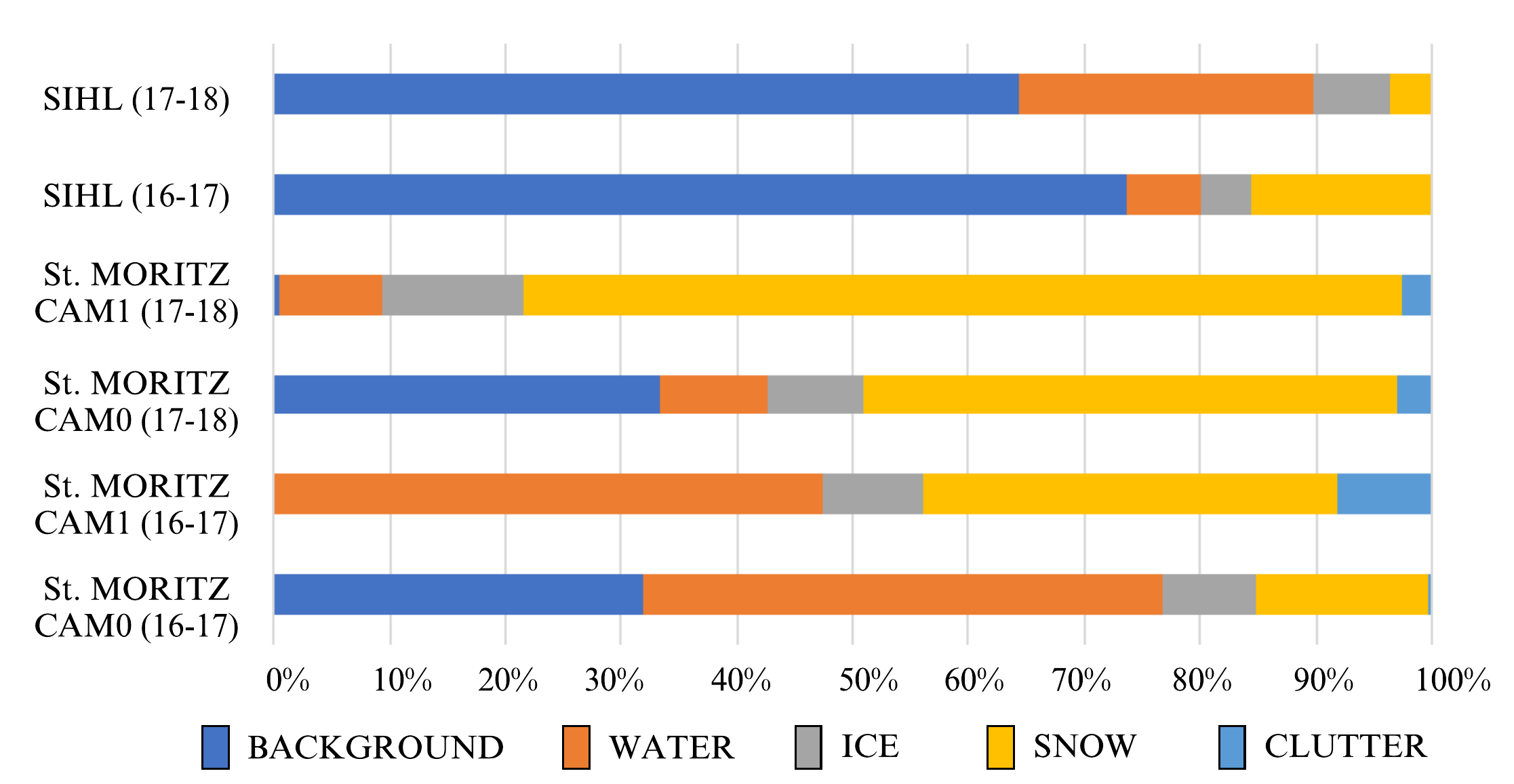}}
  \vspace{-1em}
  \caption{Class imbalance (ground-truth) in the \textit{Photi-LakeIce} dataset. Best if viewed on-screen.}
  \label{fig:classimbalance}
\end{figure}

\textbf{\textit{Photi-LakeIce} dataset.} We report lake ice
segmentation results on the \textit{Photi-LakeIce} dataset, which we
make publicly available to the research community. The dataset
comprises of images from two lakes (St.~Moritz, Sihl) and two winters
(W$2016$-$17$ and W$2017$-$18$). See Table \ref{tab:datasetlakeice} for
details. For images in this dataset, we also provide pixel-wise ground truth for
foreground-background segmentation as well as for lake ice
segmentation. There are two different, fixed webcams (Cam0 and Cam1,
see Fig.~\ref{fig:Photi-lakeice-dataset-snapshots}a and c) both
observing lake St.~Moritz at different zoom levels. The third camera
(Cam 2), at lake Sihl, rotates around one axis and observes the lake
in four different viewing directions. Example images are shown in
Fig.~\ref{fig:Photi-lakeice-dataset-snapshots}.
Additionally, Fig.~\ref{fig:classimbalance} shows the class
frequencies for all classes (background + 4 states on the lake), which
are fairly imbalanced with \textit{ice} and \textit{clutter} always
being under-represented. For lake Sihl, there are
four different camera angles involved in capturing distinct lake
views, causing the difference in background frequencies. The background frequencies of the same camera slightly vary across different winters (such as Cam0 of St. Moritz) mostly due to differences in manual annotations, as these two winters are annotated by two different operators. 
\subsection{Crowd-sourced data}
\label{section:CS_data}

As an even more extreme generalisation task than between different
webcam views, we also test the method on individual images sourced from
online image-sharing platforms. We note that there is a potential to
also include such images as complementary data sources in a monitoring
system, as long as they are time-stamped.
We employed keywords such as \textit{frozen St.~Moritz}, \textit{lake
  ice St.~Moritz}, \textit{St.~Moritz lake in winter} etc.\ to gather
lake ice images from online platforms such as
Google, Flickr, Pinterest, etc.
In total, we collected $150$ images, which are all resized to a
spatial resolution of $512\times512$ for further use. Examples are
shown in Figs.~\ref{fig:crowd2}a and \ref{fig:crowd4}a.

\section{EXPERIMENTS, RESULTS and DISCUSSION}\label{sec:experiments}
\textbf{Network details}\label{sec:trainingdetails}
All networks are implemented in Tensorflow.
The lake detection model is trained on image crops of size
$500\times500$, whereas the lake ice segmentation model, is trained with
crop of size $321\times321$.
The evaluation of the (fully convolutional) networks is always run at
full image resolution without any cropping.
The per-class losses are balanced by re-weighting the cross-entropy
loss with the inverse (relative) frequencies in the training set.
All models are trained for $100$ epochs with batch sizes of $4$ for lake
detection and $8$ for lake ice segmentation, respectively. Atrous rates are set to [$6,12,18$] in all experiments. Simple stochastic gradient
descent empirically worked better than more sophisticated optimisation
techniques. The base learning rate is set to $10^{-5}$ and reduced
according to the \textit{poly} schedule
\citep{DBLP:journals/corr/LiuRB15}.

\subsection{Results on webcam images}
\begin{table}[!ht]
\small
    \centering
    \vspace{-0.5em}
        \begin{tabular}{|l|c|c|c|c|}\hline
        \multicolumn{2}{|c|}{\textbf{Train set}} & \multicolumn{2}{c|}{\textbf{Test set}} & \multirow{2}{*}{\textbf{mIoU}}\\\cline{1-4}
    \textbf{Lakes}&\textbf{\#images}&\textbf{Lake}&\textbf{\#images}& \\
    \hline
    Silv, Moritz, Sihl& 7477&Sils&2075 & 0.93\\
    Silv, Sils, Sihl& 8456& Moritz&1096& 0.92\\
    Silv, Sils, Moritz&9104&  Sihl&448 & 0.93\\
    Silv(0,1,2), SMS&7652&Silv(3)&1900 & 0.95\\
    Silv(0,1,3), SMS&7906& Silv(2)&1646  & 0.95\\
    Silv(0,2,3), SMS&8676& Silv(1) &876 & 0.90\\
    Silv(1,2,3), SMS&8041& Silv(0) &1511 & 0.94\\\hline
\end{tabular}
\caption{Mean IoU (mIoU) values of the leave-one-cam-out experiments
  for lake detection. Silv(0,1,2,3) are the different camera angles
  for lake Silvaplana. SMS refers to \{Sils, St.~Moritz,
    Sihl\}.}
\label{tab:lakedetection}
\end{table}
\normalsize

\begin{figure}
 \centering
  \includegraphics[height=3.5cm,width=0.85\columnwidth]{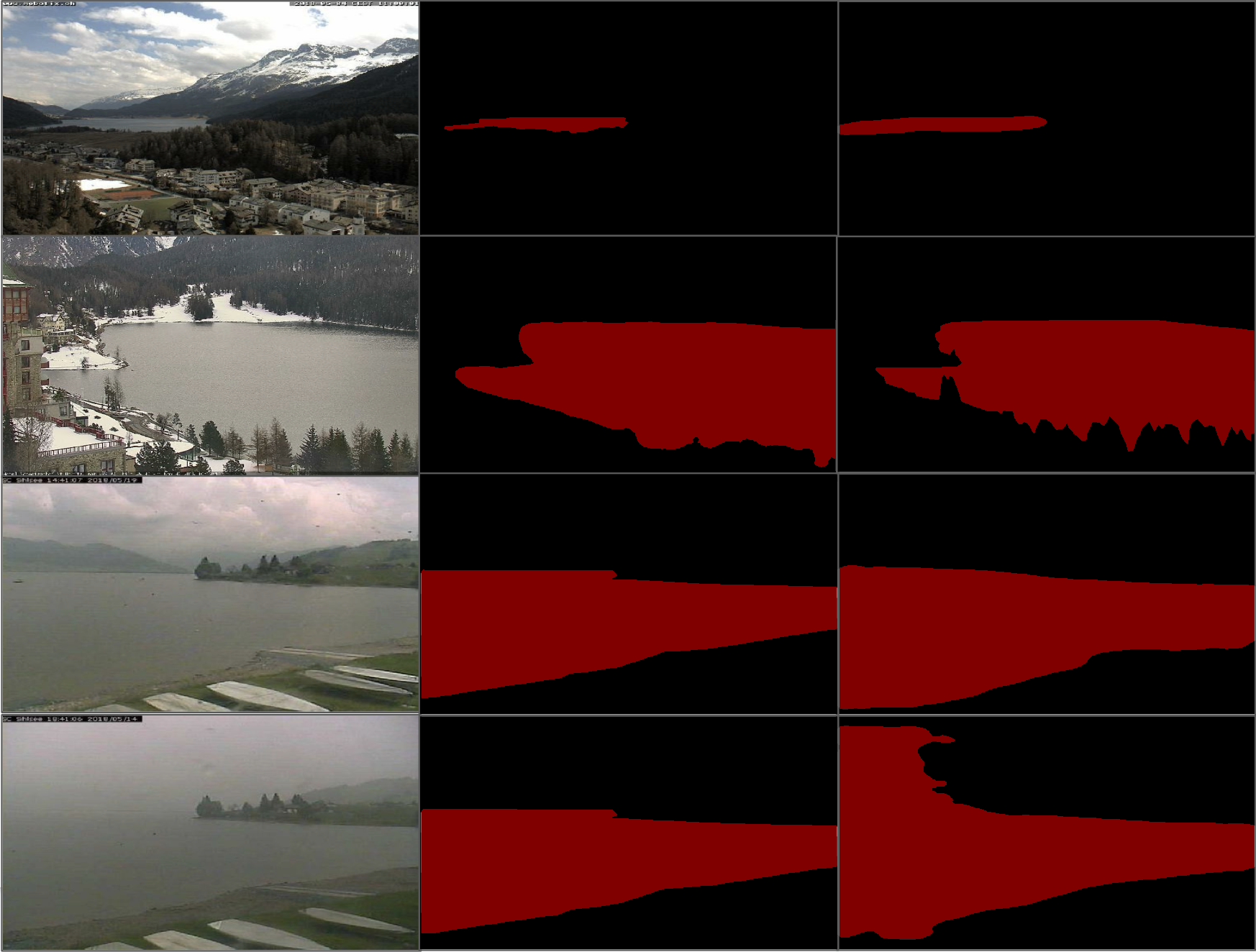}\\
  \subfigure[Image]{\hspace{.25\linewidth}}
  \subfigure[Ground truth]{\hspace{.30\linewidth}}
  \subfigure[Prediction]{\hspace{.30\linewidth}}
  \vspace{-0.75em}
  \caption{Results of lake detection using \textit{Deeplab v3+}. The
    first three rows shows successful cases, a failure case is
    displayed in the last row.}
  \label{fig:lakedetection}
\end{figure}
%
    

\begin{figure}[ht]
  \small
  \subfigure[Cam0 (test) results when trained using the data from Cam0 (train).]{\includegraphics[height=5cm,width=4cm]{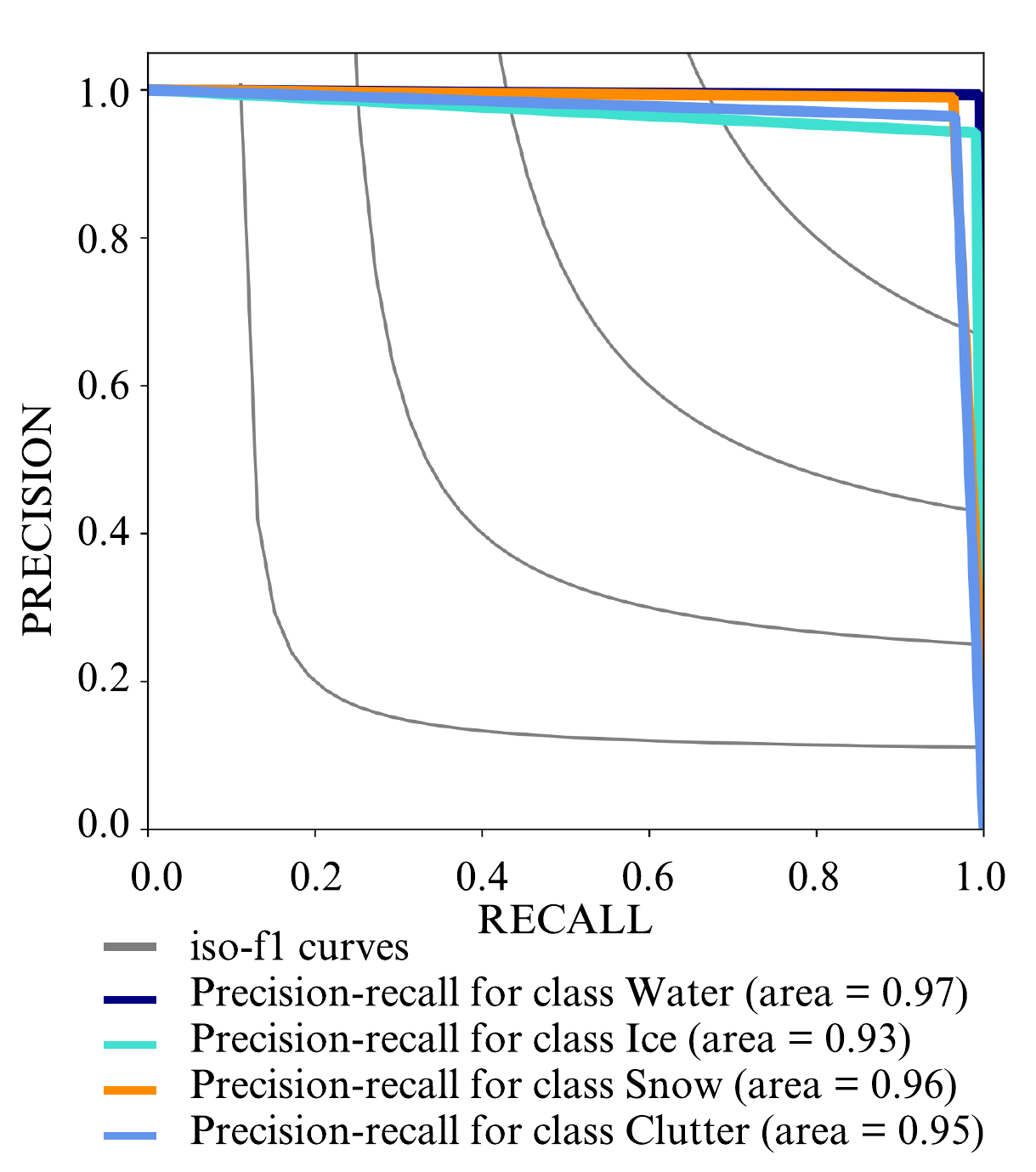}}
  \subfigure[Cam1 (test) results when trained using the data from Cam1 (train).]{\includegraphics[height=5cm,width=4cm]{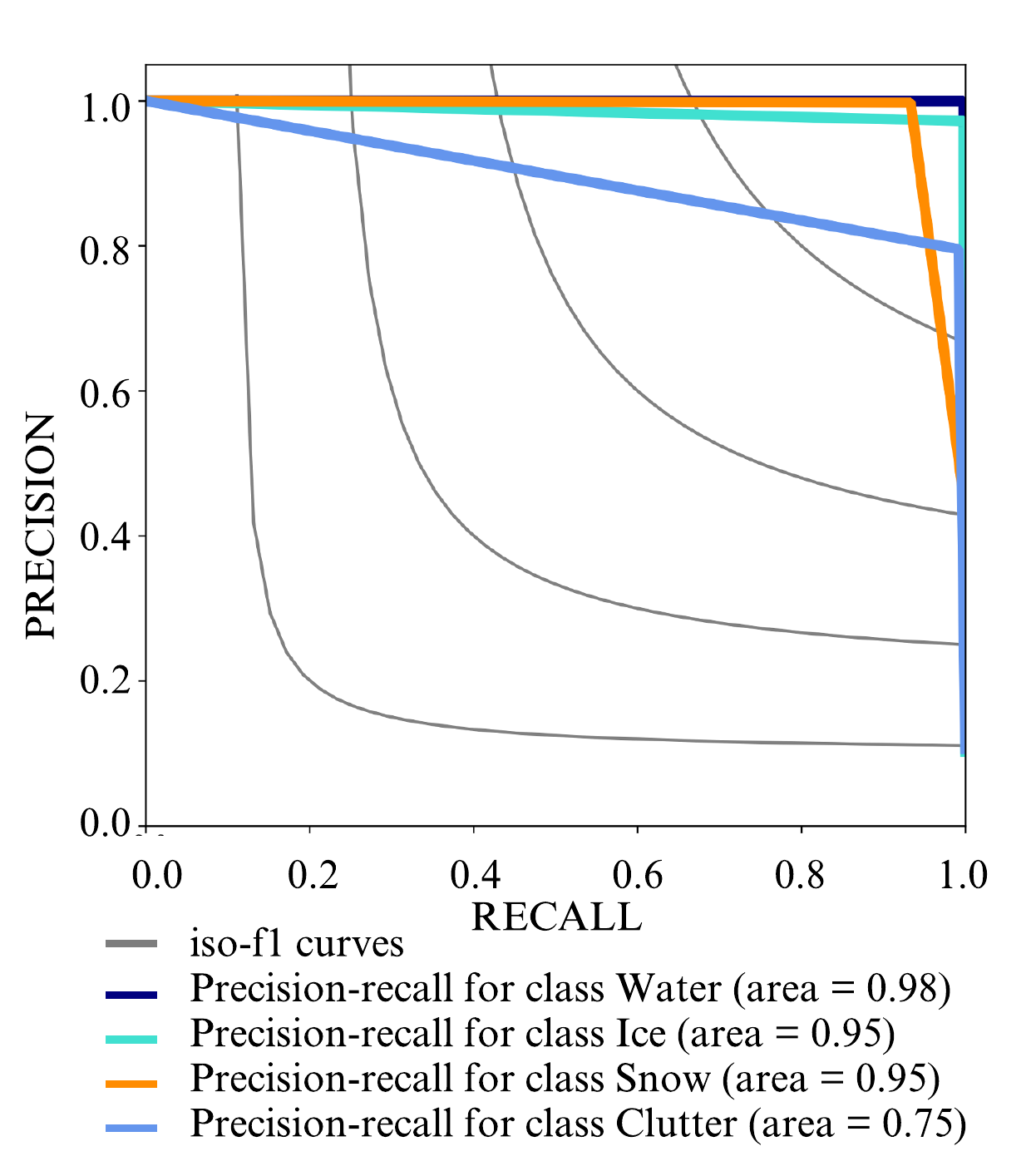}}\\\vspace{-1em}
  \subfigure[Cam0 results when trained using the data from Cam1.]{\includegraphics[height=5.1cm,width=4cm]{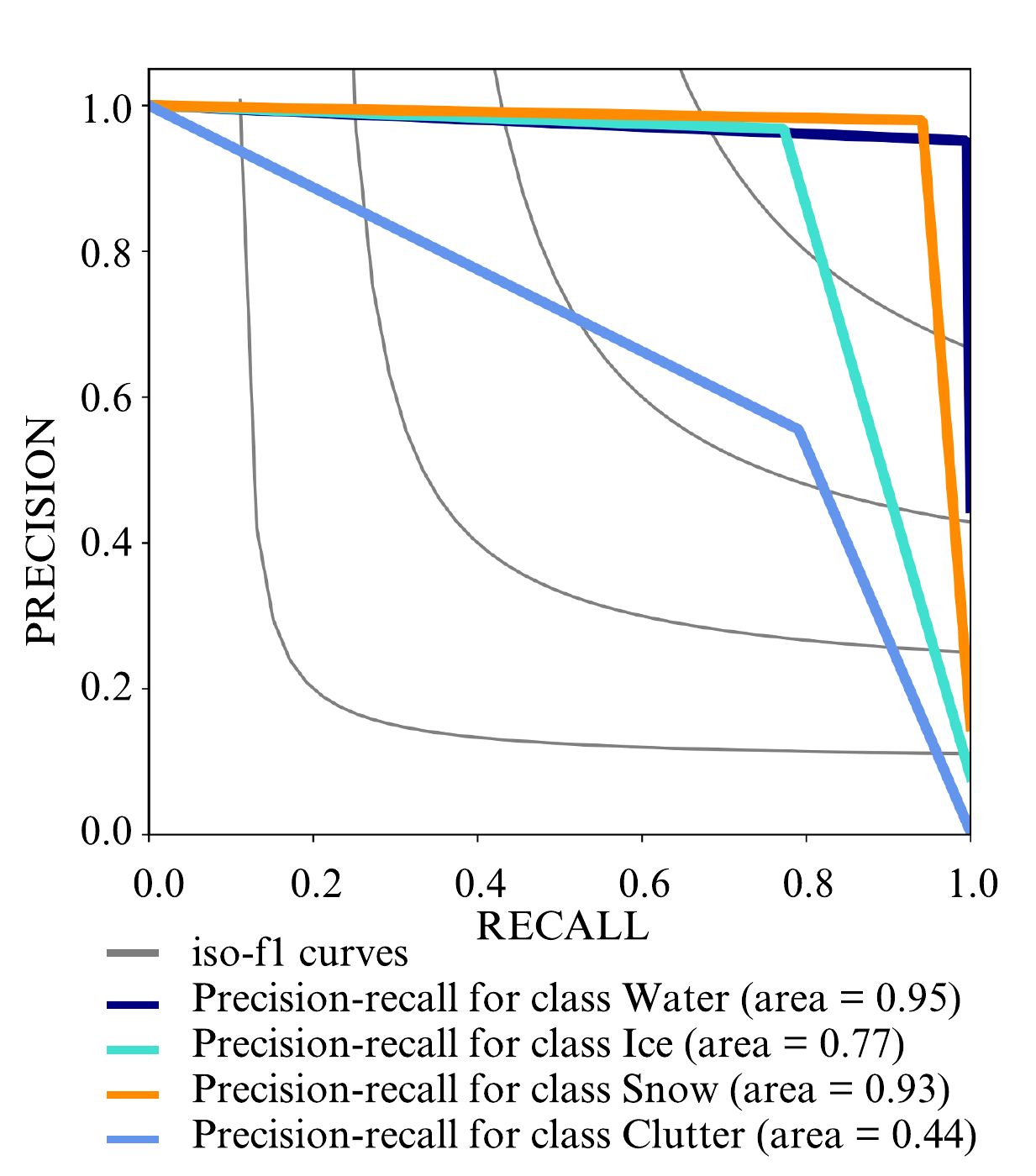}}
  \subfigure[Cam1 results when trained using the data from Cam0.]{\includegraphics[height=5cm,width=4cm]{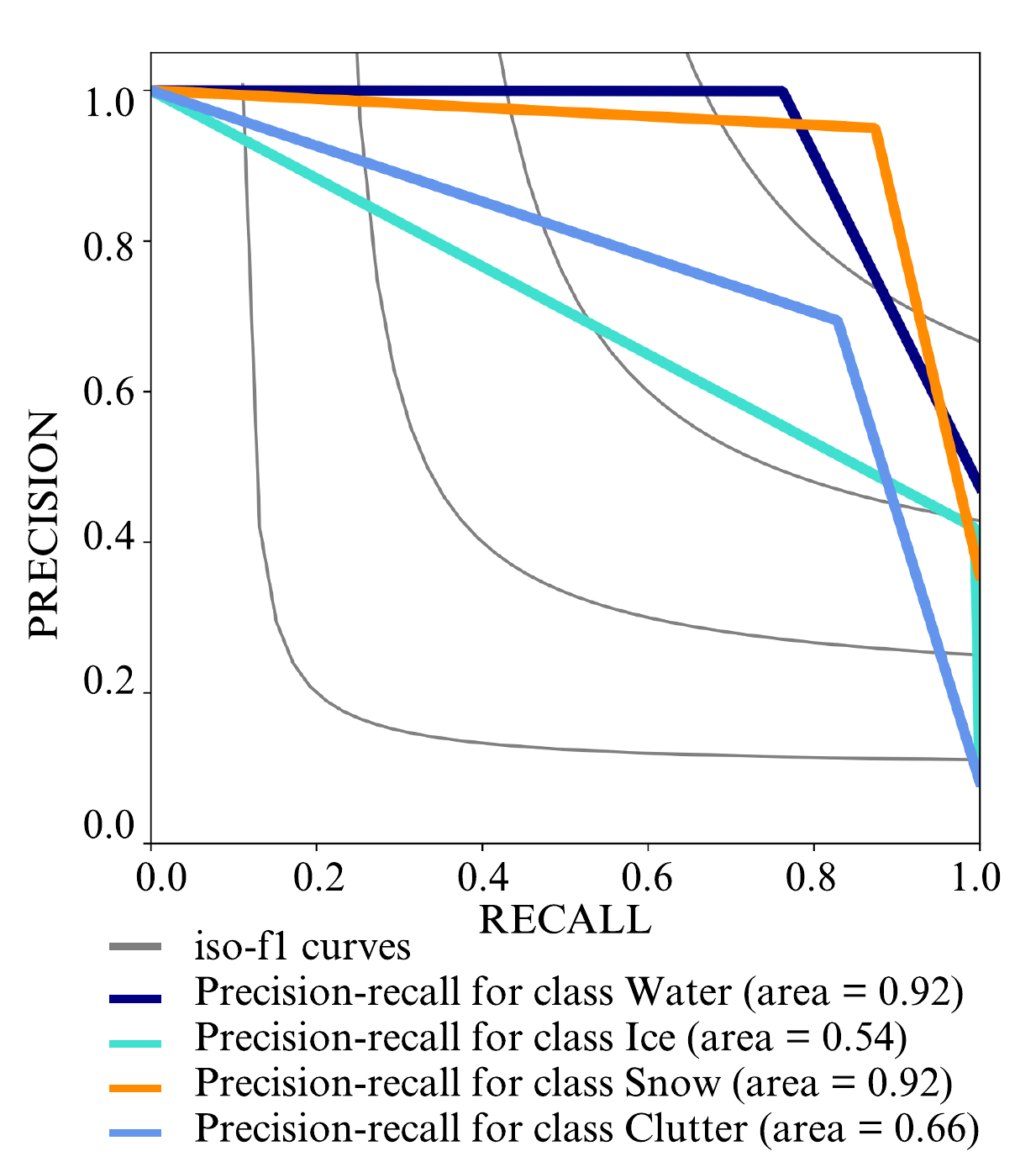}}\\\vspace{-1em}
  
  \caption{Precision-recall curves (Lake St.~Moritz). Best if viewed on screen.} 
\label{fig:figure_placement}
\end{figure}

\normalsize
\begin{figure*}[ht]
 \centering
  \includegraphics[height=1.5cm,width=0.85\columnwidth]{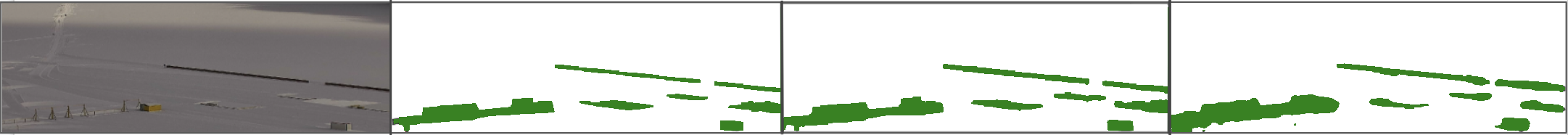}
  \subfigure[Webcam image]{\hspace{.20\linewidth}}
  \subfigure[Ground truth]{\hspace{.20\linewidth}}
  \subfigure[Deep-U-Lab]{\hspace{.20\linewidth}}
  \subfigure[Deeplab v3+]{\hspace{.20\linewidth}}
  \vspace{-1em}
  \caption{\textit{Deeplab v3+} vs.\ \textit{Deep-U-lab}. Segmentation
    boundaries are visibly crisper and more accurate with additional
    skip connections.}
  \label{fig:withskips}
\end{figure*}
\begin{table*}[ht]
\small
    \centering
        \begin{tabular}{|l|c|c|c|c|l|l|l|l|l|}\hline
   \multirow{2}{*}{\textbf{Lake}} & \multicolumn{2}{|c|}{\textbf{Train set}} & \multicolumn{2}{|c|}{\textbf{Test set}} & \multirow{2}{*}{\textbf{Water}} & \multirow{2}{*}{\textbf{Ice}} & \multirow{2}{*}{\textbf{Snow}} & \multirow{2}{*}{\textbf{Clutter}} & \multirow{2}{*}{\textbf{mIoU}}\\\cline{2-5}
   & \textbf{Cam} & \textbf{Winter} & \textbf{Cam} & \textbf{Winter} & &  &  &  &  \\\hline
   St.~Moritz & Cam0 & 16-17 & Cam0 & 16-17 & 0.98 / \textcolor{gray}{0.70} & 0.95 / \textcolor{gray}{0.87} & 0.95 / \textcolor{gray}{0.89} & 0.97 / \textcolor{gray}{0.63} & 0.96 / \textcolor{gray}{0.77} \\
   St.~Moritz & Cam1 & 16-17 & Cam1 & 16-17 & 0.99 / \textcolor{gray}{0.90} & 0.96 / \textcolor{gray}{0.92} & 0.95 / \textcolor{gray}{0.94} & 0.79 / \textcolor{gray}{0.62} & 0.92 / \textcolor{gray}{0.85} \\
   St.~Moritz & Cam0 & 17-18 & Cam0 & 17-18 & 0.97 & 0.88 & 0.96 & 0.87 & 0.93 \\
   St.~Moritz & Cam1 & 17-18 & Cam1 & 17-18 & 0.93 & 0.84 & 0.92 & 0.84 & 0.89 \\
   Sihl & Cam2 & 16-17 & Cam2 & 16-17 & 0.79 & 0.62 & 0.81 & --- & 0.74 \\
   Sihl & Cam2 & 17-18 & Cam2 & 17-18 & 0.81 & 0.69 & 0.86 & --- & 0.79 \\
   \hline
   St.~Moritz & Cam0 & 16-17 & Cam1 & 16-17 & 0.76 / \textcolor{gray}{0.36} & 0.75 / \textcolor{gray}{0.57} & 0.84 / \textcolor{gray}{0.37} & 0.61 / \textcolor{gray}{0.27} & 0.74 / \textcolor{gray}{0.39}  \\
   St.~Moritz & Cam1 & 16-17 & Cam0 & 16-17 & 0.94 / \textcolor{gray}{0.32} & 0.75 / \textcolor{gray}{0.41} & 0.92 / \textcolor{gray}{0.33} & 0.48 / \textcolor{gray}{0.43} & 0.77 / \textcolor{gray}{0.37} \\
   St.~Moritz & Cam0 & 17-18 & Cam1 & 17-18 & 0.62 & 0.66 & 0.89 & 0.42 & 0.64  \\
   St.~Moritz & Cam1 & 17-18 & Cam0 & 17-18 & 0.59 & 0.67 & 0.91 & 0.51 & 0.67 \\
   \hline
   St.~Moritz & Cam0 & 16-17 & Cam0 & 17-18 & 0.64 / \textcolor{gray}{0.45} & 0.58 / \textcolor{gray}{0.44} & 0.87 / \textcolor{gray}{0.83} & 0.59 / \textcolor{gray}{0.40} & 0.67 / \textcolor{gray}{0.53}  \\
   St.~Moritz & Cam0 & 17-18 & Cam0 & 16-17 & 0.98  &0.91 & 0.94 &0.58  &0.87   \\
   St.~Moritz & Cam1 & 16-17 & Cam1 & 17-18 & 0.86 / \textcolor{gray}{0.80} & 0.71 / \textcolor{gray}{0.58} & 0.93 / \textcolor{gray}{0.92} & 0.57 / \textcolor{gray}{0.33} & 0.77 / \textcolor{gray}{0.57}  \\
   St.~Moritz & Cam1 & 17-18 & Cam1 & 16-17 & 0.93 & 0.76 & 0.86 & 0.65 & 0.80  \\
   Sihl & Cam2 & 16-17 & Cam2 & 17-18 & 0.61 & 0.14 & 0.35 & --- &0.51 \\
   Sihl & Cam2 & 17-18 & Cam2 & 16-17 & 0.41 & 0.18 & 0.45 & --- & 0.50 \\
   \hline
    \end{tabular}
    \caption{Lake ice segmentation results (IoU) on the
      \textit{Photi-LakeIce} dataset. For comparison, we also show
      results of \cite{prs_report} where available, in
      \textcolor{gray}{grey}. We outperform them in all instances.}
\label{tab:moritz0and1winter1617}
\end{table*}
\normalsize

\textbf{Lake detection results.}
\label{sec:quantlakedetection}
Only summer images are used to avoid problems due to snow cover (on
both the lake and the surroundings). The model performed well with
$\geq$0.9 mean Intersection-over-Union (mIoU) score (weighted according
to the class distribution in the train set) in all
cases, see Table~\ref{tab:lakedetection}.
We are not aware of any previous work on lake detection in webcam
images, but note that water bodies are in general segmented rather
well in RGB images.
Figure~\ref{fig:lakedetection} shows the qualitative results of the lake
detection, including a failure case in the last row. It can be
observed that the wrong classification occurs in a rather foggy image
that is difficult to judge even for humans.
Also, note the fairly good prediction in the first row, a challenging
case where the lake covers $<$5\% of the image.

\textbf{Lake ice segmentation results.}
\label{sec:quantlakeicesegmentation}

Table \ref{tab:moritz0and1winter1617} shows the results for lake ice
segmentation on the \textit{Photi-LakeIce} dataset. Exhaustive
experiments were performed to evaluate \textit{same camera train-test}
(rows 1-6), \textit{cross-camera train-test} (rows 7-10) and
\textit{cross-winter train-test} (rows 11-16).

For the \textit{same camera train-test} experiments, the model is
trained randomly on $75\%$ of the images and tested on the remaining
$25\%$. As shown in Table \ref{tab:moritz0and1winter1617} (rows 1 and
2), the mIoU scores of the proposed approach are respectively $19$ and $7$ percent points higher than the ones reported by
\cite{prs_report}.
For lake St.~Moritz, in addition to the results on the winter
$2016$-$17$, we report results for the winter $2017$-$18$. Additionally,
we present results on a second, more challenging lake (Sihl), for both
winters.
As can be seen in Fig.~\ref{fig:Photi-lakeice-dataset-snapshots}, the
images from lake Sihl (Cam2) are of significantly lower quality, with
severe compression artifacts, low spatial resolution, and small lake
area in pixels, which amplifies the influence of small, miss-classified
regions on the error metrics. Consequently, our method performs worse
than for St.~Moritz, but still reaches $>$76\% correct classification
under the rather strict IoU metric.
We note that there is no \textit{clutter} class since no events take
place on lake Sihl.

%
%

The main drawback of prior studies on lake ice detection is their
models' inability to generalise from one camera view to
another~\citep{muyan_lakeice,prs_report}.
For the \textit{cross-camera} experiments (rows 7-10, Table
\ref{tab:moritz0and1winter1617}), our model is trained on all images
from one camera and tested on all images from another camera. As per
Table \ref{tab:moritz0and1winter1617}, for winter $2016$-$17$, the mIoU
results for that experiment surpass the \textit{FC-DenseNet}
\citep{prs_report} by margin of $35$ to $40$ percent points.
This huge improvement clearly shows the superior ability of the
deep learning architecture to learn generally applicable
``visual concepts'' and avoid overfitting to specific sensor
characteristics and viewpoints. For completeness, we also report
\textit{cross camera} results for winter $2017$-$18$. They are a bit
worse than those for $2016$-$17$, due to more complex appearance and
lighting during that season (e.g., black ice) that cause increased
confusion between ice and water.

For an operational system, the ultimate goal is to train on the data
from a set of lakes from one, or a few, winters and then apply the
system in further winters, without the need to annotate further
reference data.
Hence, we also performed \textit{cross-winter} experiments to assess
the generalisation across winters. i.e., the model is trained on the
data from one full winter and tested on the data acquired from the
same viewpoint over a second winter.
The results (Table \ref{tab:moritz0and1winter1617}, rows 11-16) show
that the model also generalises quite well across winters. For
St.~Moritz, a model trained on winter $2016$-$17$ reaches an IoU of
$77$\% on $2017$-$18$, a gain of 20 percent points over prior art \citep{prs_report}. For
Cam0, there is also a substantial gain of $14$ percent points.
%
%
It can, however, also be seen that there is still room for improvement
in less favorable imaging settings such as lake Sihl, where the
segmentation of ice and snow in a different winter largely fails.

%
\begin{figure}[ht]
 \centering
  \includegraphics[height=2.5cm,width=0.85\columnwidth]{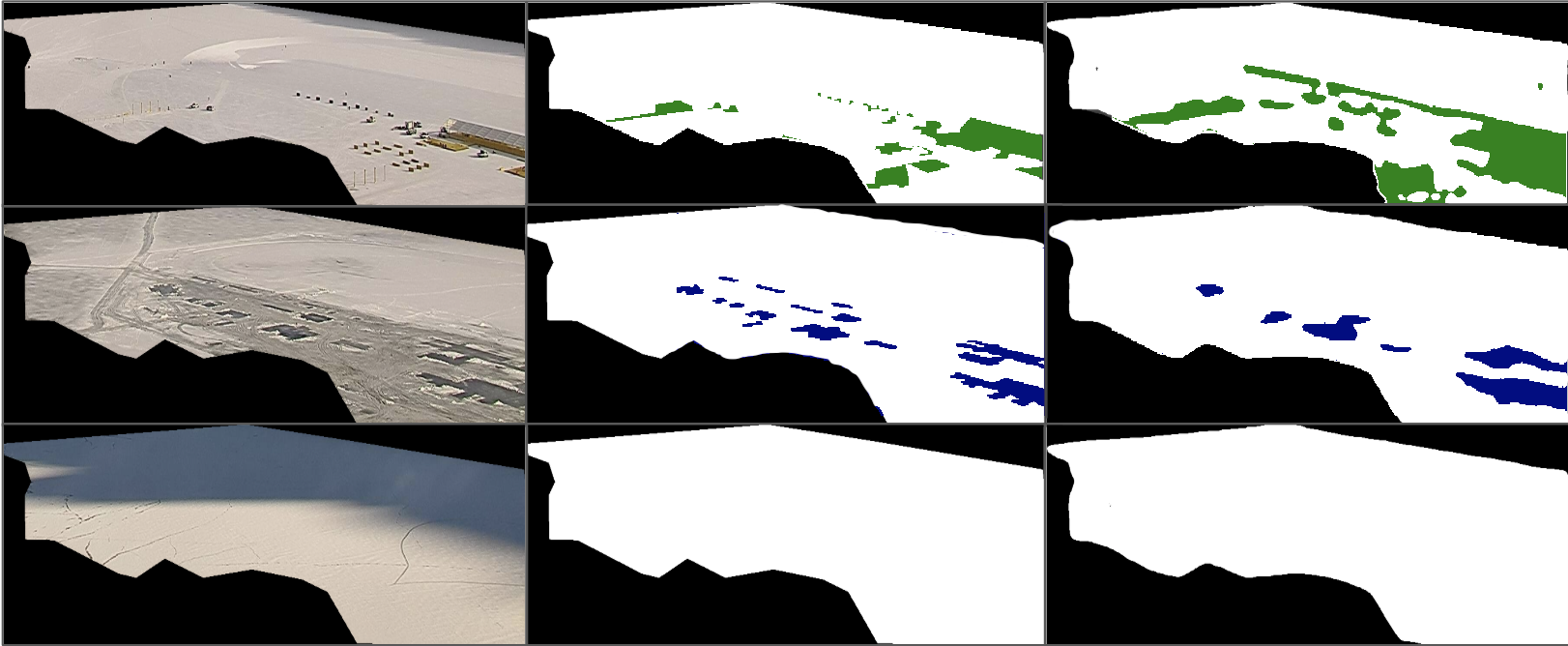}
  \subfigure[Image]{\hspace{.25\linewidth}}
  \subfigure[Ground Truth]{\hspace{.30\linewidth}}
  \subfigure[Prediction]{\hspace{.30\linewidth}}
  \vspace{-1em}
  \caption{Cam0 results when the model is trained only on Cam1.}
  \label{fig:cam1on0}
\end{figure}
\begin{figure}[ht]
 \centering
  \includegraphics[height=2cm,width=0.85\columnwidth]{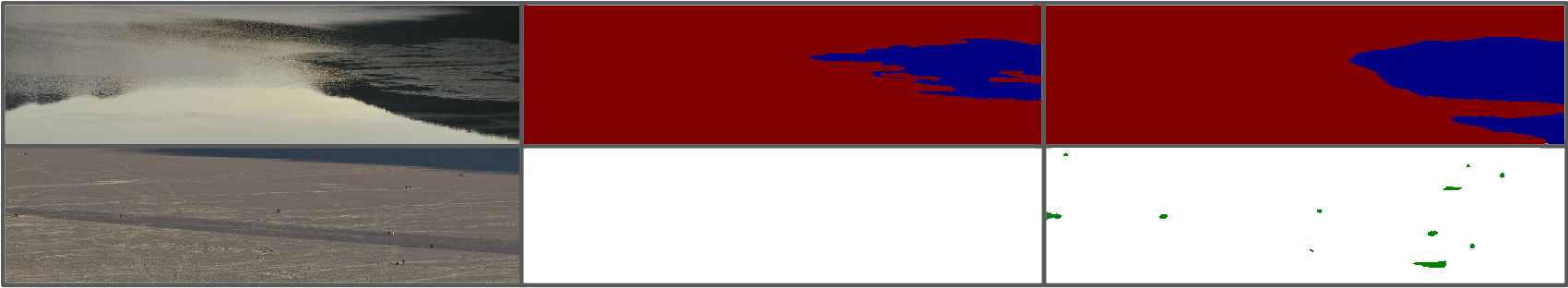}
  \subfigure[Image]{\hspace{.25\linewidth}}
  \subfigure[Ground Truth]{\hspace{.30\linewidth}}
  \subfigure[Prediction]{\hspace{.30\linewidth}}
  \vspace{-1em}
  \caption{Cam1 results when the model is trained only on Cam0.}
  \label{fig:cam0on1}
\end{figure}
\begin{figure*}[ht]
\begin{center}
   \includegraphics[height=2cm,width=17cm]{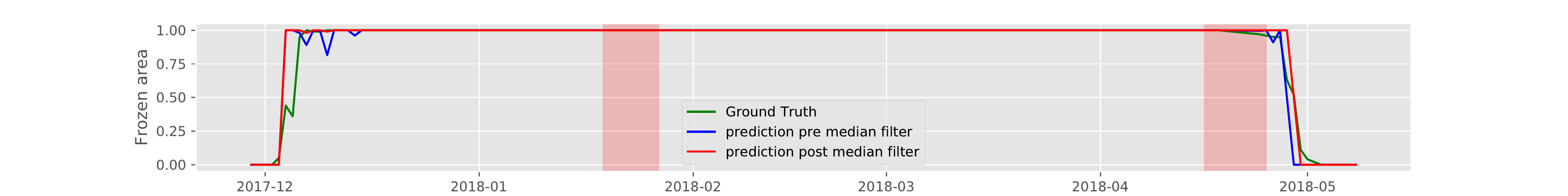}\vspace{-1em}
   \caption{Frozen area time series with- and without
      post-processing: results of Cam0 when the network is trained
      using the data from Cam1. Red bars indicate periods of data
      gaps, where no images are stored due to technical failures.}
\label{fig:icecoverageon0}
\end{center}
\end{figure*}
%
%

For a more comprehensive assessment of the per-class results we also
generate precision-recall curves, see
Fig.~\ref{fig:figure_placement}. It can be seen that the performance
for \textit{ice} and \textit{clutter} is inferior to the other two
classes. A large part of the errors for \textit{clutter} are actually
due to imprecise \emph{ground truth} rather than prediction errors of
the model, as the annotated masks for thin and intricate structures
like flagpoles, food stalls and individual people on the lake tend to
be ``bulk annotations'' that greatly inflate the (relative) amount of
clutter in the ground truth, leading to large (relative) errors.
According to the curves, thresholds of 0.60 precision and 0.80 recall shows good a trade-off between the true-positive and false-positive rates for \textit{cross-camera results}. However for \textit{same-camera results}, the thresholds are much better ranging from 0.80 for Cam1 to 0.90 for Cam0.

Qualitative example results are shown in Figs.~\ref{fig:cam1on0} and
\ref{fig:cam0on1}. Sometimes the images are even confusing for humans
to annotate correctly, e.g., Fig.~\ref{fig:cam1on0}, row 2 shows an
example of ice with smudged snow on top, for which the ``correct''
labeling is not well-defined.
We note that our segmentation method is robust against cloud/mountain
shadows cast on the lake (row 3). In another interesting case
(Fig.~\ref{fig:cam0on1}, row 2) the network ``corrects'' human
labeling errors, where humans are present on the frozen lake, but not
annotated due to their small size.

\textbf{Ice-on/off results.}
\label{sec:iceonoff}
Freeze-up and break-up periods are of particular interest for climate
monitoring.
\begin{table}[ht]
  \centering
  \small
    \begin{tabular}{|l|c|c|c|}
    \hline
        \textbf{Lake} & \textbf{Winter} & \textbf{Ice-on} &  \textbf{Ice-off} \\
        \hline
   St.~Moritz& 16-17 & 14.12.16 & 18.03-26.04.17 \\
   St.~Moritz& 17-18 & 06.12.17 & 27.04.18 \\
   Sihl&16-17& 29.12.16,& 31.12.17, \\
   &  & 04.01.17, & 05.01.17,\\
   &  & 07.01.17,  &  11.01.17, \\
   && 10.02.17 & 14.02.17 \\
   Sihl& 17-18 & 29.12.17, & 02.01.18,\\
   && 15.02.18, &  23.03.18, \\
   & & 27.03.18,& 05.04.18, \\
   &&  11.04.18 & 16.04.18\\
   \hline
\end{tabular}
\caption{Ice-on/off dates predicted by our approach.}
\label{tab:iceonofftable}
\end{table}
\normalsize
%
To estimate the \textit{ice-on/off} dates, we produce a daily time
series of the (fractional) frozen lake area in a camera's field of
view, for the winter $2017-18$ (Figs. \ref{fig:icecoverageon0}). The areas in individual images are
aggregated with a daily median, then smoothed with another 3-day
median.
The latter filters out individual days with difficult conditions and
improves the model predictions by almost $3\%$. The estimated
\textit{ice-on/off} dates are shown in Table
\ref{tab:iceonofftable}. 
%
We determined the \textit{ice-on/off} dates for lake St.~Moritz from
Cam0, which covers a larger portion of the lake.
For lake Sihl, multiple \textit{ice-on} and \textit{off} dates are
found, as that lake is in a warmer (lower) region of Switzerland and
froze/thawed four times within the same winter. See Table
\ref{tab:iceonofftable}.

\begin{table}[ht]
\small
  \centering
    \begin{tabular}{|l|c|c|c|c|}\hline
            \textbf{Water} & \textbf{Ice} & \textbf{Snow} & \textbf{Clutter} & \textbf{mIoU} \\\hline
          0.60 & 0.32 & 0.71 & 0.79 & 0.60 \\\hline
\end{tabular}
\caption{Lake ice segmentation results (IoU) on crowd-sourced dataset.}
\label{tab:crowd3}
\end{table}
\normalsize
\begin{figure}[ht]
 \centering
  \includegraphics[height=2.5cm,width=0.85\textwidth]{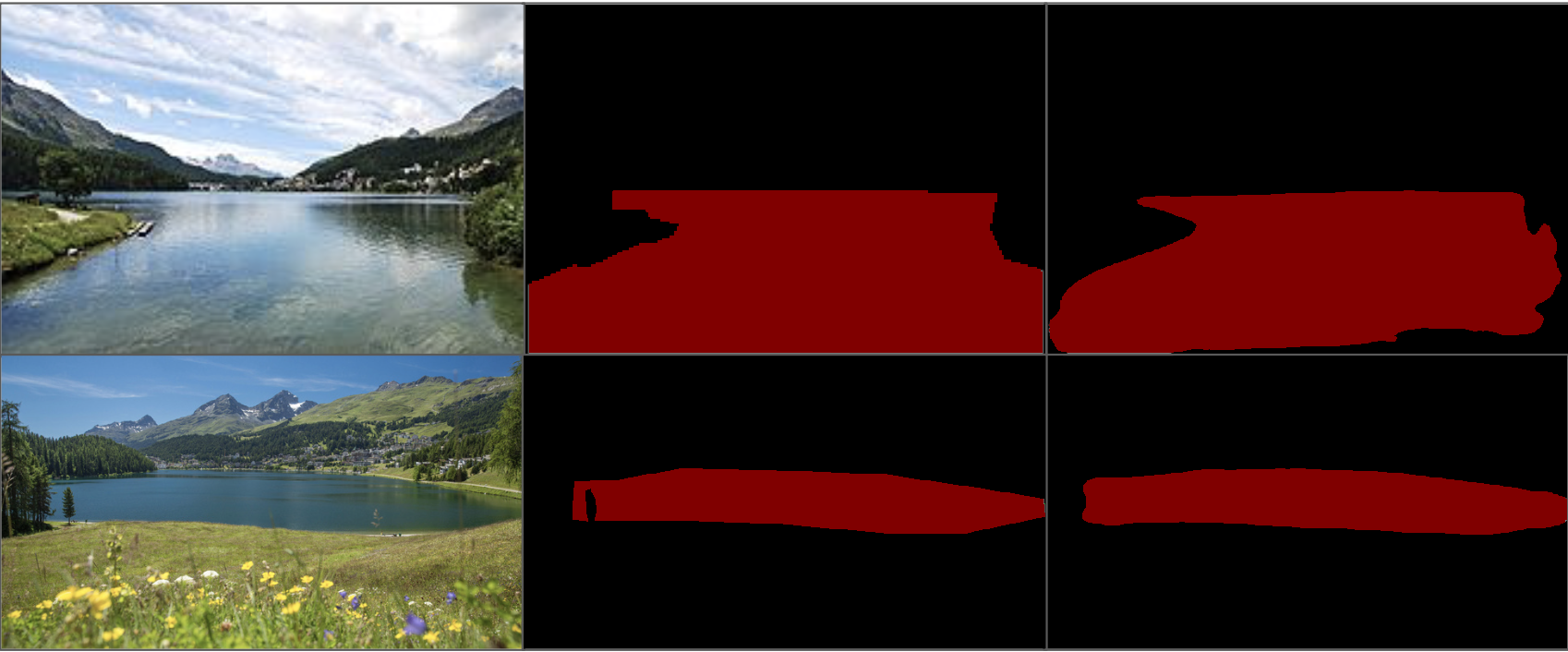}
  \subfigure[Image]{\hspace{.25\linewidth}}
  \subfigure[Ground truth]{\hspace{.30\linewidth}}
  \subfigure[Prediction]{\hspace{.30\linewidth}}
  \vspace{-1em}
  \caption{Lake detection on crowd-sourced data.}
  \label{fig:crowd2}
\end{figure}
\vspace{0.5em}
\subsection{Results on crowd-sourced images}

Crowd-sourced images have a rather different data distribution, among
others due to better image sensors and optical components, less
aggressive compression, more vivid colours due to on-device
electronics and image editing, etc.
Thus, they are, arguably, an even more challenging test of model
generalisation. With the model trained on webcam images (St.~Moritz winter $2016$-$17$), lake
detection in crowd-sourced images yields an IoU of 75\% for the
background and 64\% for the lake.
Qualitative results are
shown in Fig. \ref{fig:crowd2}.

\begin{figure}[ht]
 \centering
  \includegraphics[height=2.5cm,width=0.85\textwidth]{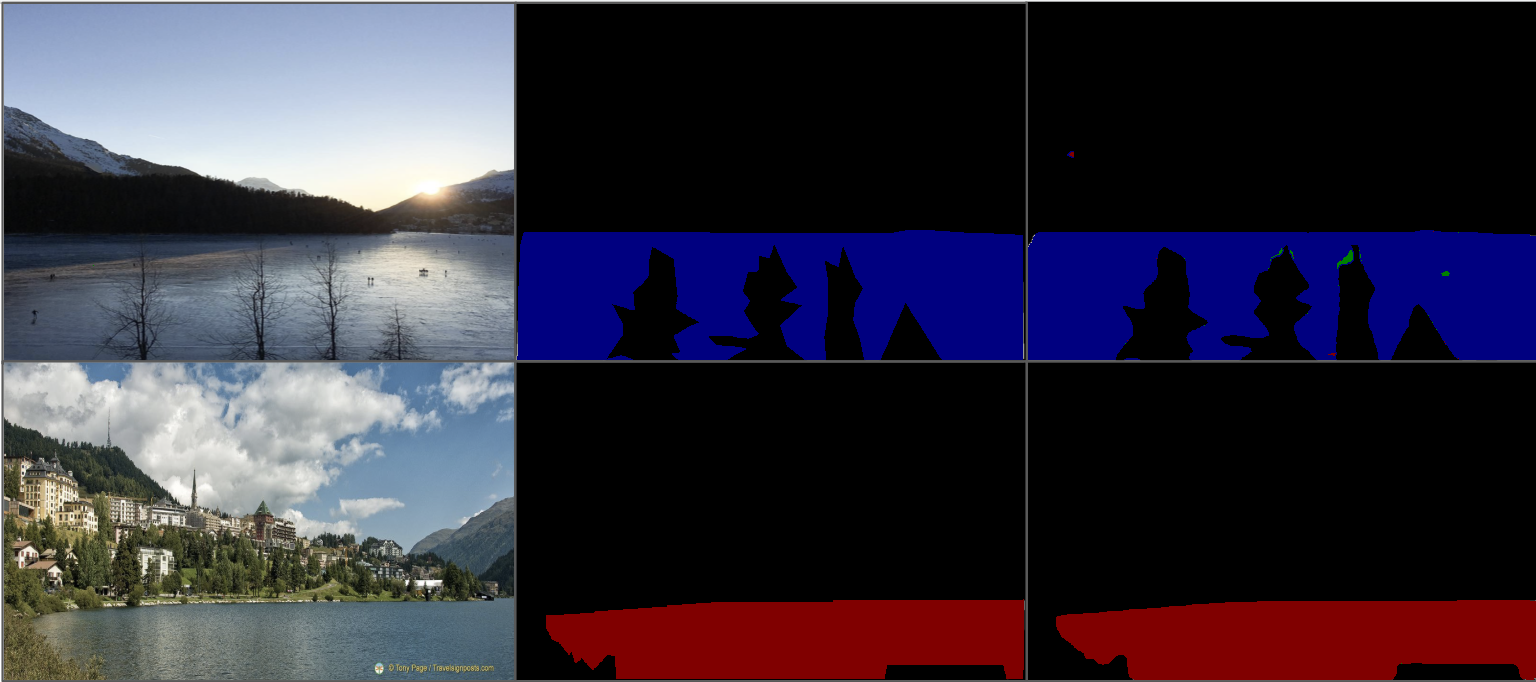}
  \subfigure[Image]{\hspace{.30\linewidth}}
  \subfigure[Ground truth]{\hspace{.30\linewidth}}
  \subfigure[Prediction]{\hspace{.30\linewidth}}
  \vspace{-1em}
  \caption{Lake ice segmentation on crowd-sourced data.}
  \label{fig:crowd4}
\end{figure}

For the semantic segmentation task, we apply the model trained on
webcam images (St.~Moritz winter $2016$-$17$) on the crowd-sourced images. Quantitative
results are presented in Table \ref{tab:crowd3}. Note that these are
still significantly better than the cross-camera generalisation
results of \cite{prs_report}. Qualitative
examples are shown in Fig.~\ref{fig:crowd4}.
\subsection{Discussion}

A natural question that arises is: Why does \textit{Deep-U-Lab}
perform a lot better compared to \textit{FC-DenseNet} for lake ice
detection? While it is difficult to conclusively attribute the empirical performance of deep neural networks to specific architectural choices, we speculate that there are two main reasons why Deep-U-Lab is superior to FC-DenseNet.

First, by following a popular ``standard'' architecture, we can start from very well pre-trained weights -- yet another confirmation that the benefits of pre-training on big datasets often outweigh the perceived domain gap to specific sensor and application settings. Unfortunately, we could not complete the comparison by training Deep-U-Lab from scratch with our data, as this did not converge.
Second, our model has a much larger receptive field around every pixel, due to the atrous convolutions. It appears that long-range context and texture, which only our model can exploit, play an important role for lake ice detection.


\section{CONCLUSION AND OUTLOOK}\label{sec:conclusion}

One conclusion that we drew from our study is that the previous,
pioneering attempts \citep{muyan_lakeice,prs_report} underestimated the
potential of deep convolutional networks for lake ice detection with
webcams.
We found that with modern high-performance architectures like
\textit{Deeplab v3+}, in particular our variant \textit{Deep-U-lab},
segmentation results are near-perfect within the data of one camera
over one winter (i.e., in the scenario where a portion of the data is
annotated manually, then extrapolation to the remaining frames is
automatic).
Moreover, also generalisation to different views of the same lake, as
well as to different winters with the same camera viewpoint, works
fairly well. Especially the latter case is very interesting for an
operational scenario: it is quite likely that a system trained on data
from two or three winters would reach well above 80\% IoU for all
classes of interest.
Moreover, it appears within reach to even complement dedicated
monitoring cameras (or, in touristic places, public webcams) with
amateur images opportunistically gleaned from the web.

An open question for future work is how to minimise the initial
annotation effort, to simplify the introduction of monitoring systems
especially at new locations. A fascinating extension could be to adopt
ideas from few-shot learning and/or active learning to quickly adapt
the system to new locations.


%
\vspace{-0.5em}
\section*{ACKNOWLEDGEMENTS}\label{ACKNOWLEDGEMENTS}
This work is part of the project ``Integrated lake ice monitoring and generation of sustainable, reliable, long time series'', funded by Swiss Federal Office of Meteorology and Climatology \emph{MeteoSwiss} in the framework of GCOS Switzerland. This work was partially funded by the Sofja Kovalevskaja Award of the Humboldt Foundation. We thank Muyan Xiao, Konstantinos Fokeas and Tianyu Wu for their help with labelling the data.
\vspace{-0.5em}
{
	\begin{spacing}{1.17}
		\normalsize
		\bibliography{authors} 
	\end{spacing}
}

\end{document}